\pdfoutput=1

\documentclass[11pt]{article}

\usepackage{ACL2023}

\usepackage{times}
\usepackage{latexsym}
\usepackage{algorithm}
\usepackage[noend]{algpseudocode}
\usepackage{multirow}
\usepackage{multicol}
\usepackage{booktabs}
\usepackage{tabularx}
\usepackage{enumitem}
\usepackage{xcolor}
\newcommand\crule[3][black]{\textcolor{#1}{\rule{#2}{#3}}}
\definecolor{fuchsia}{rgb}{1.0, 0.0, 1.0}
\definecolor{purple(html/css)}{rgb}{0.5, 0.0, 0.5}
\definecolor{softpink}{rgb}{0.89, 0.47, 0.76}
\usepackage[caption=false]{subfig}
\usepackage{amsmath}

\usepackage[T1]{fontenc}

\usepackage[utf8]{inputenc}

\usepackage{microtype}

\usepackage{inconsolata}

\usepackage{booktabs}
\usepackage[disable]{todonotes}

\makeatletter
\newcommand*\iftodonotes{\if@todonotes@disabled\expandafter\@secondoftwo\else\expandafter\@firstoftwo\fi}  %
\makeatother

\newcommand{\note}[4][]{\todo[author=#2,color=#3,size=\scriptsize,fancyline,caption={},#1]{#4}} %

\newcommand{\anuj}[2][]{\note[#1]{anuj}{cyan!40}{#2}}

\usepackage{caption}
\setlist{noitemsep}

\usepackage{float}

\title{When to Use Efficient Self Attention? \\ Profiling Text, Speech and Image Transformer Variants}

\author{Anuj Diwan, Eunsol Choi, David Harwath \\ Department of Computer Science \\ The University of Texas at Austin \\ \texttt{\{anuj.diwan, eunsol, harwath\}@utexas.edu}}

\begin{document}

\maketitle

\begin{abstract}
We present the first unified study of the efficiency of self-attention-based Transformer variants spanning text, speech and vision. We identify input length thresholds (\textit{tipping points}) at which efficient Transformer variants become more efficient than vanilla models, using a variety of efficiency metrics (latency, throughput, and memory). To conduct this analysis for speech, we introduce L-HuBERT, a novel local-attention variant of a self-supervised speech model. We observe that these thresholds are (a) much higher than typical dataset sequence lengths and (b) dependent on the metric and modality, showing that choosing the right model depends on modality, task type (long-form vs. typical context) and resource constraints (time vs. memory). By visualising the breakdown of the computational costs for transformer components, we also show that non-self-attention components exhibit significant computational costs. We release our profiling toolkit at \url{https://github.com/ajd12342/profiling-transformers}.

\end{abstract}

\section{Introduction and Related Work}

Transformers~\cite{vaswani2017attention} are widely adopted across NLP~\citep{devlin-etal-2019-bert, brown2020gpt3}, Speech Processing~\citep{Mohamed_2022} and Computer Vision~\cite{dosovitskiy2020vit}. Studies have shown that scaling models up improves performance~\citep{palm}, making efficiency an important research topic. Many Transformer variants focus on improving the efficiency of self-attention, motivated by its asymptotic quadratic time/space complexity with respect to the input sequence length.\footnote{We refer the readers to ~\citet{tay2020effsurvey} for a comprehensive overview of efficient Transformers.} While these Transformer variants are designed be asymptotically faster, in practice they may actually be slower, especially given modest input lengths that are typical of many tasks.

Our paper presents two main analyses. First, we visualize the \textit{layerwise} efficiency of such models to locate bottlenecks and attempt to answer the question \textit{``is self-attention the true bottleneck?''} We find that in the non-asymptotic case, non-self-attention layers contribute significantly to the overall cost, especially for speech architectures due to the input waveform tokenizer in models like HuBERT~\cite{Hsu2021HuBERTSS}. Second, \textit{when} should we use self-attention-based efficient Transformers? Comparing efficient variants with vanilla models at different input lengths, we find that this \textit{tipping point} where efficient variants outperform vanilla architectures is much higher than typical input lengths of existing benchmarks across all modalities, calling into question the efficacy of using such efficient Transformers and requiring new benchmarks. We introduce a local-attention variant of a speech Transformer, HuBERT, to conduct this analysis. Together, our analyses suggest that current approaches that focus on improving self-attention might not be the most effective for improving efficiency.

\section{Efficiency Metrics} \label{sec:profilingmethods}
Model efficiency is an umbrella term for a suite of \textit{efficiency metrics}, which do not always correlate with, and sometimes contradict, each other~\cite{effmisnomer2021}. Further, different metrics are relevant to different end use-cases. To cover most use-cases, we evaluate a set of four metrics; two for computational time and two for memory usage:

\noindent \textbf{Throughput}: Number of examples processed per sec, given inputs of a given sequence length, using the maximum possible batch size for a given GPU.  %

\noindent \textbf{Latency-Inference}: Time (in ms) to run inference for $1$ unbatched input of a given sequence length. %

\noindent \textbf{Max-Memory}: The allocated GPU memory (MiB) for processing $1$ input of a given sequence length.

\noindent \textbf{Parameter Count:} Number of model parameters.

We profile models in all modalities in \textit{training} mode and \textit{inference} mode. For training, while Transformer architectures often use prediction heads with a larger output space (e.g., for text generation), we choose a lightweight binary classification head for profiling. %

\paragraph{Layerwise Efficiency Metrics}
We also profile some metrics and models in a \textbf{layerwise} fashion to locate their efficiency bottlenecks. Our goal is twofold: a) provide an empirical approach to efficient model design, as an alternative to theoretical analyses or mental models (e.g. self-attention is $O(n^2)$) and b) empirically answer the question \textit{"to what degree is self-attention the bottleneck?"}

We identify important layer types (Self-Attention, Feedforward, etc.\@) and profile the Latency-Inference and Parameter Count metrics per-layer-type to obtain a fine-grained understanding of which layer types and indices (layer $0$ vs $11$) contribute the most to model efficiency costs. We use param counts as a proxy for memory (profiling real layerwise memory usage is non-trivial due to Pytorch memory allocation intricacies).
We profile the layers depicted in Figure~\ref{fig:diagram}; more details in Appendix~\ref{app:layerdescriptions}.
\begin{figure}
    \centering
    \includegraphics[width=0.8\linewidth]{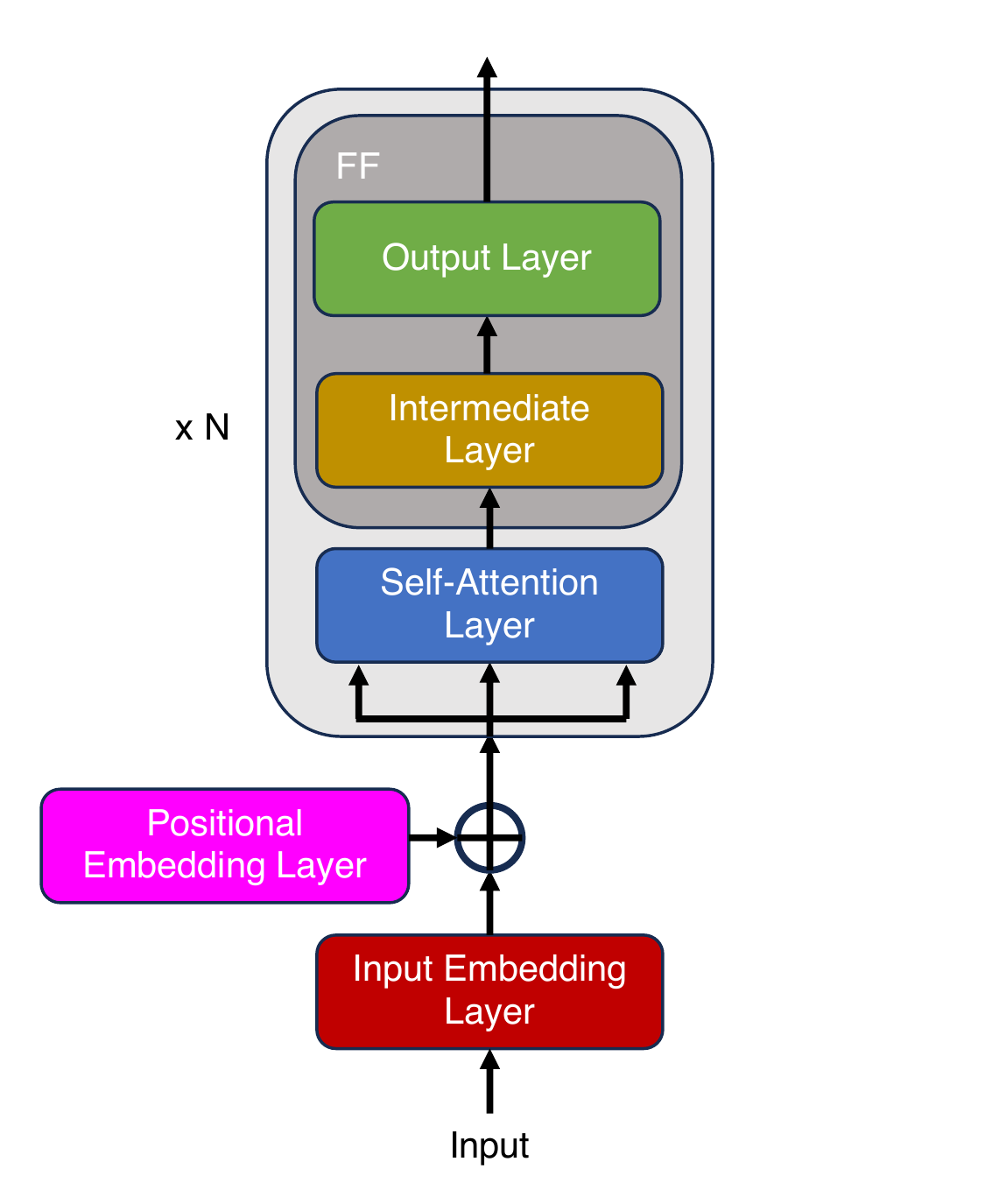}
    \caption{Transformer layer types profiled in our layerwise efficiency profiling experiments.}\vspace{-1em}
    \label{fig:diagram}
\end{figure}

\anuj{Moved layer descriptions to appendix}

\section{Local-Attention Speech Model}
\label{sec:localhubertdesc}

Efficient transformers~\cite{xiong2021nystromformer,Ma2021LunaLU} have not received as much attention in Speech as they have in NLP and CV, perhaps due to two reasons. First, there is a relative lack of long-context speech benchmarks as compared to those in NLP (LRA~\cite{tay2021long} and QuALITY~\cite{Pang2021QuALITYQA}). Second, when performing speech tasks like automatic speech recognition (ASR), it is typical to segment a long speech signal into small individual utterances and perform ASR independently on each. For example, most Librispeech examples are less than 5 seconds. Many popular speech models like HuBERT~\cite{Hsu2021HuBERTSS} tokenize the waveform at 50 tokens per second, implying that a typical utterance has only several hundred tokens; far below the number of tokens in long-context NLP tasks. Nevertheless, textless speech models~\cite{lakhotia-etal-2021-generative} are becoming more feasible, motivating the modelling of long speech utterances. %

\begin{table}
    \centering
    \small
    \setlength{\tabcolsep}{0.5em}
    \begin{tabular}{ccc}
        \toprule
        \textbf{Model} & \textbf{WER} $\downarrow$ & \textbf{WER (w/ FT)} $\downarrow$ \\
        \midrule
        HuBERT Base & 7.09 & 3.4\\
        L-HuBERT (32 | 100) & 21.06 | 14.48 & 8.52 | 7.39 \\
        \bottomrule
    \end{tabular}
    \caption{WERs on the SUPERB ASR task.}
    \label{tab:local_hubert_asr}
    \vspace{-15pt}
\end{table}

\paragraph{Local HuBERT Model}
To investigate the efficiency of the self-attention layer in speech models, we introduce the \textit{Local HuBERT} model which replaces HuBERT's self-attention with the Longformer~\cite{Beltagy2020Longformer} sliding-window self-attention. In this attention mechanism, every token attends to tokens within a local window context, rather than the full token sequence.
Our model is similar to the temporally windowed-attention Transformer acoustic model proposed by~\citet{alastruey2021} for speech translation; our approach differs by using the self-supervised HuBERT model as our basis, and we evaluate on ASR. Choosing the appropriate window size for the local attention context is key; we explore 32 and 100 token contexts, corresponding to 640 ms and 2 s, inspired by phone recognition models that typically incorporate similar context sizes~\cite{Peddinti2015ATD,Yeh2019TransformerTransducerES}.

\paragraph{ASR Results}
We initialize the L-HuBERT model with pretrained HuBERT Base weights (pretrained with full self-attention), and then replace self-attention with sliding-window self-attention; due to limited compute, we did not pretrain L-HuBERT from scratch using sliding-window attention. We then evaluate L-HuBERT on Librispeech~\cite{librispeech} ASR via the SUPERB~\cite{superb} benchmark under two settings; a) \textbf{Freeze}: freezing the model and only training projection weights and b) \textbf{Finetune}: fully finetune the model. We use the default S3PRL\footnote{\url{https://github.com/s3prl/s3prl}} hyperparams; but we train for 200k steps for Freeze and 104k steps for Finetune. Both models converge by 104k steps; we train Freeze for longer to eke out as much performance as possible, while we stop training Finetune due to limited compute.

We report Word Error Rate (WER) on Librispeech \texttt{test-clean} in Table~\ref{tab:local_hubert_asr}; lower is better. In the frozen setting (middle column), we see a large WER increase over HuBERT; we hypothesize that this is due to the attention layer mismatch since we initialize L-HuBERT with HuBERT weights that were pretrained with full self attention, rather than pretraining L-HuBERT from scratch. However, in the finetuning setting, the gap between HuBERT Base and L-HuBERT narrows considerably and using a larger window size achieves better performance. As our L-HuBERT model is a reasonable architecture capable of moderate ASR performance, we can continue to study its computational efficiency (we profile the window-100 variant).

\section{Methods and Implementation}
We analyze the Base versions of the BERT~\cite{devlin-etal-2019-bert}, Longformer~\cite{Beltagy2020Longformer} and Nyströmformer~\cite{xiong2021nystromformer} models for text; the HuBERT~\cite{Hsu2021HuBERTSS} and L-HuBERT (Section~\ref{sec:localhubertdesc}) models for speech; and Vision Transformer~\cite{dosovitskiy2020vit} and Swin Transformer~\cite{liu2021Swin} models for vision; BERT, HuBERT and ViT are standard Transformer encoder architectures. Longformer, L-HuBERT and Swin use fixed-pattern self-attention while Nyströmformer uses approximate self-attention.

\begin{table}
    \centering
    \small
    \setlength{\tabcolsep}{0.23em}
    \begin{tabular}{lrrrrrr}
    \toprule
    Model & Emb & Pos & SA & Interm & Output & Others \\ \midrule
    BERT   & 23.8M & - & 29M & 28.3M & 28.3M & 0.6M  \\ 
    HuBERT & 4.2M & 5.1M & 29M & 28.3M & 28.3M & 0.2M\\ 
    ViT & 0.6M & - & 29M & 28.3M & 28.3M & 0.6M  \\ \bottomrule
    \end{tabular}
    \caption{Layer-wise parameter counts. Emb: Input Embedding, Pos: Positional Emb. SA: Self-Attention, Interm: Intermediate.}
    \label{tab:layer_wise_param}
    \vspace{-10pt}
\end{table}

\subsection{Sequence Length Ranges}
\label{subsec:inputrange}
\begin{table*}[]
    \centering
    \small
    \setlength{\tabcolsep}{0.5em}
    \begin{tabular}{c|ccccccc|cccccc}
    \toprule
    & \multicolumn{7}{c|}{Text} & \multicolumn{6}{c}{Speech} \\
    \midrule
    Dataset & SST & MNLI & SQ & ON & CNN & HPQA & TQA & TEDL & LJS & VoxC & Libri & S-SQuAD & Spotify\\
    \# of tokens & $23$ & $36$ & $177$ & $506$ & $863$ & $1,316$ & $6,589$ & $301$ & $328$ & $390$ & $615$ & $3080$ & $101400$  \\
    \bottomrule
    \end{tabular}\vspace{-0.5em}
    \caption{Average token sequence lengths. Left to right: Stanford Sentiment Treebank, MultiNLI, SQuAD2.0, OntoNotes, CNN-DailyMail, HotpotQA, TriviaQA, TEDLIUM, LJSpeech, VoxCeleb Speaker Recognition, Librispeech, Spoken SQuAD, Spotify Podcasts.}
    \label{tab:inputranges}
    
\end{table*}
We profile our models on a wide range of input sequence lengths to cover both avg. sequence lengths of commonly used contemporary datasets (Table~\ref{tab:inputranges}) and typical sequence lengths of long-context tasks. Details about how we compute sequence lengths in Table~\ref{tab:inputranges} can be found in Appendix~\ref{app:tokenlengths}. Most image datasets use images resized to $224$ or $512$ pixels. Below, $\texttt{range}(a,b,c)$ means a range from $a$ to $b$ in steps of $c$. Since there is no difference between synthetic and real inputs from a computational complexity standpoint, we use synthetic inputs to more easily control for their sequence lengths.

\noindent \textbf{Text Modality} \quad
The input is \textit{`This is a sentence.'} repeated $n$ times, $n \in \texttt{range}(10,560,10)$ i.e. $\texttt{range}(62,3362,60)$ tokens for all tokenizers. %

\noindent \textbf{Speech Modality} \quad The inputs have durations in $\texttt{range}(1,50,0.5)$ sec i.e. $\texttt{range}(50,2500,25)$ tokens for all featurizers (CNNs with $20$ ms framerate). Our sampling strategy is in Appendix~\ref{app:speechinputrange}.

\noindent \textbf{Image Modality} \quad We use square inputs of dimension in $\texttt{range}(32,1024,32)$ pixels by rescaling a fixed image. The \# tokens depend on featurizer patch size, which is different for different models.

\subsection{Implementational Details}
We profile time-based metrics (latency/throughput) using Pytorch CUDA Events\footnote{\url{https://pytorch.org/docs/stable/generated/torch.cuda.Event.html}} by executing 20 iterations sequentially. The first few iterations serve as GPU warm-start; thus, we report the average of the last 10. We record Max-Memory with \texttt{torch.cuda.max\_memory\_allocated()} and param counts with \texttt{torchinfo}~\cite{torchinfo}.

To profile throughput, we \textit{approximate} the max batch size that fits on a single GPU using a linear estimator; more details in Appendix~\ref{app:throughput}. Finally, we profile the layerwise Latency-Inference metric using \texttt{torchprof}~\cite{awwong1-torchprof}. We attach profiling hooks to modules of interest (e.g. Self-Attention, Embedding), giving us execution times of their \texttt{forward()} functions (other modules/functions are not profiled).
We use the Huggingface~\cite{wolf-etal-2020-transformers} implementations of text and image models and \texttt{fairseq}~\cite{ott-etal-2019-fairseq} implementations for speech models;
more details in Appendix~\ref{app:implmodels}.

\section{Profiling Results}
\label{sec:results}
\subsection{Layerwise Profiling Results}
\label{sec:layerwiseanalysis}
\begin{table*}[htb!]
    \small
    \centering
    \begin{tabular}{c}
    \crule[red]{0.8em}{0.8em} Input Embedding \quad \crule[fuchsia]{0.8em}{0.8em} Positional Embedding \quad \crule[blue]{0.8em}{0.8em} Self-Attention \quad \crule[yellow]{0.8em}{0.8em} Intermediate \quad \crule[green]{0.8em}{0.8em} Output \quad 
    \crule[black]{0.8em}{0.8em} Other \\
    \end{tabular}
    \begin{tabular}{ccc}
    Text (BERT) & Speech (HuBERT) & Vision (ViT) \\
    \includegraphics[width=0.3\textwidth]{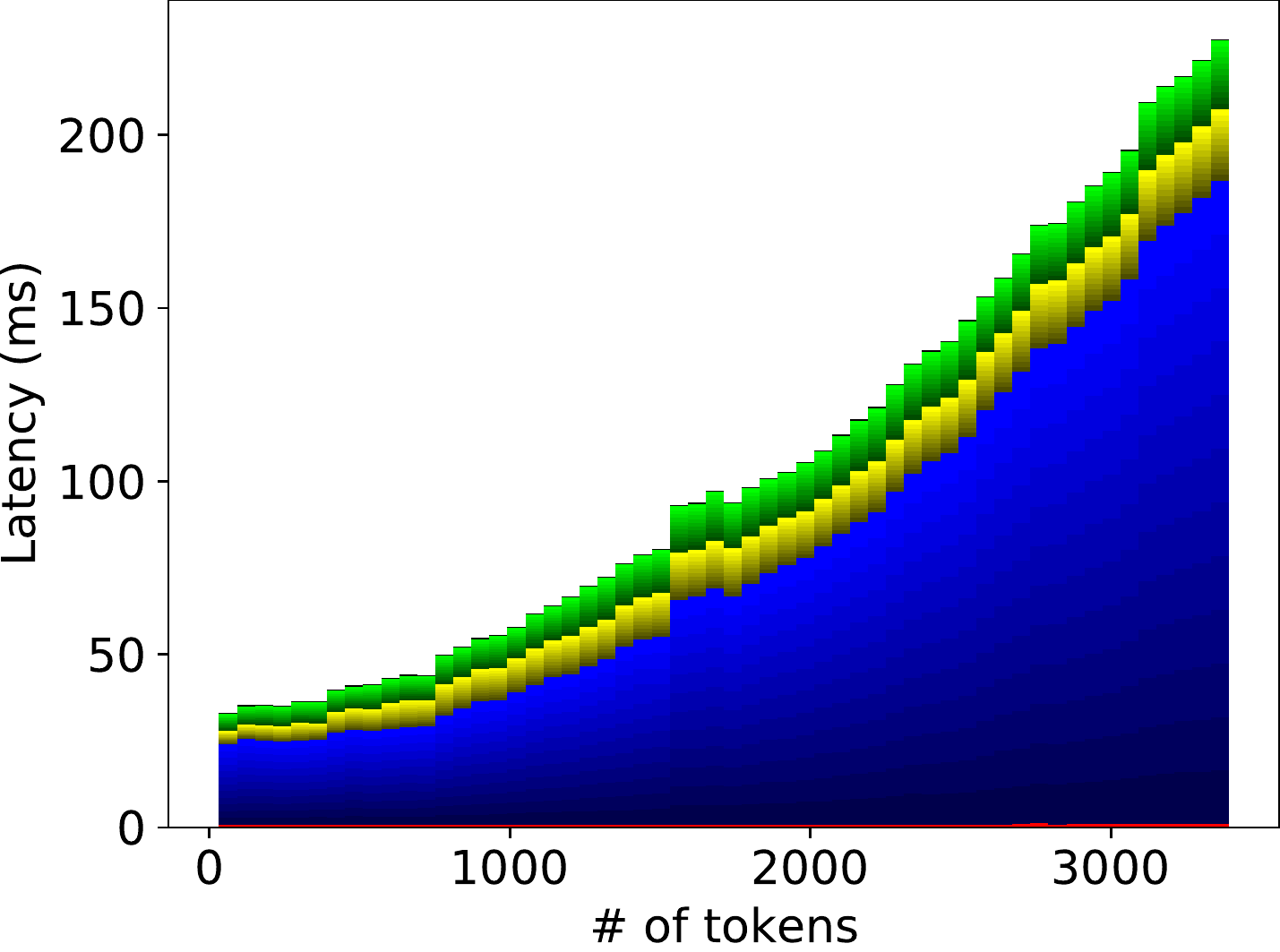} & \includegraphics[width=0.3\textwidth]{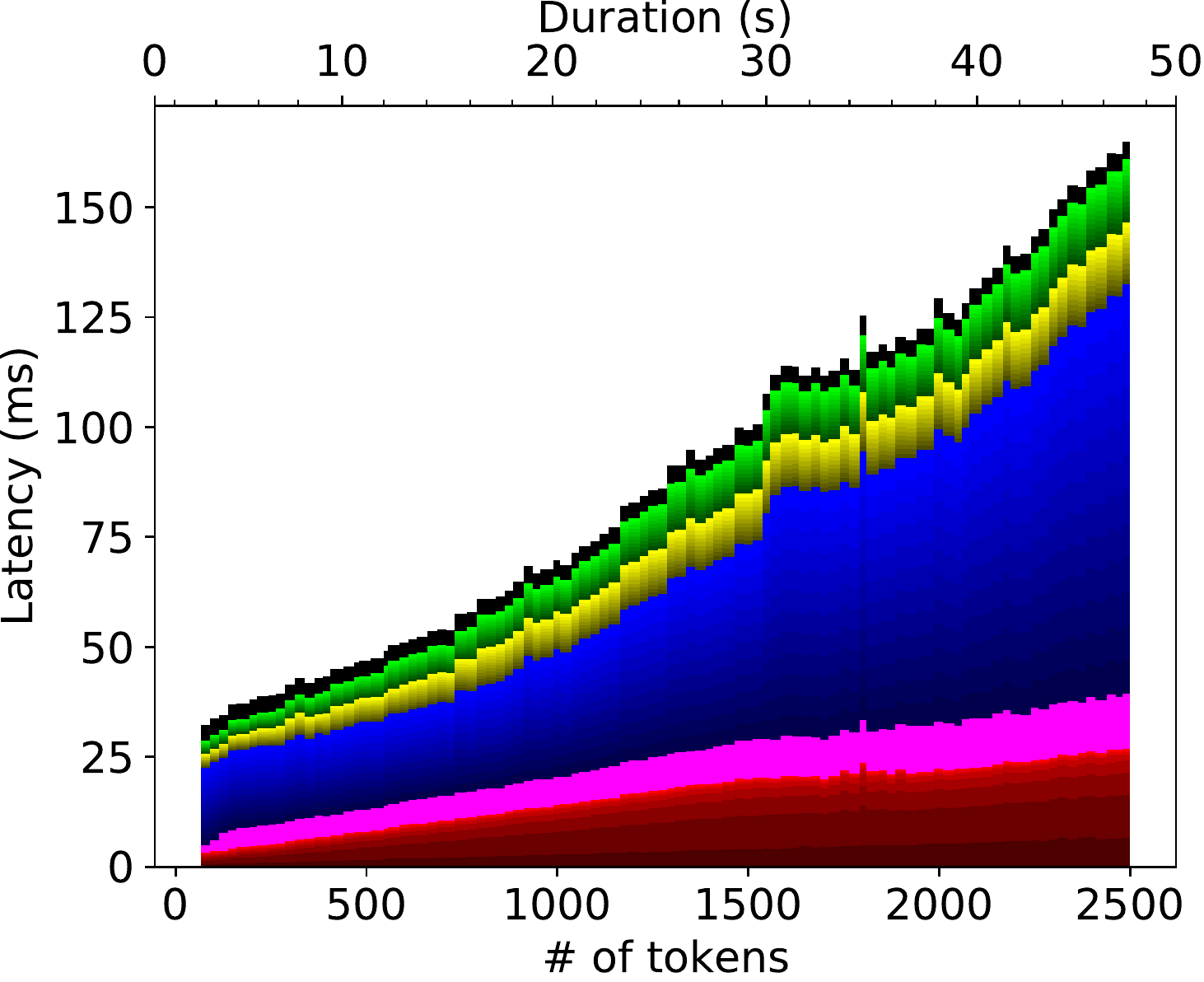} & \includegraphics[width=0.3\textwidth]{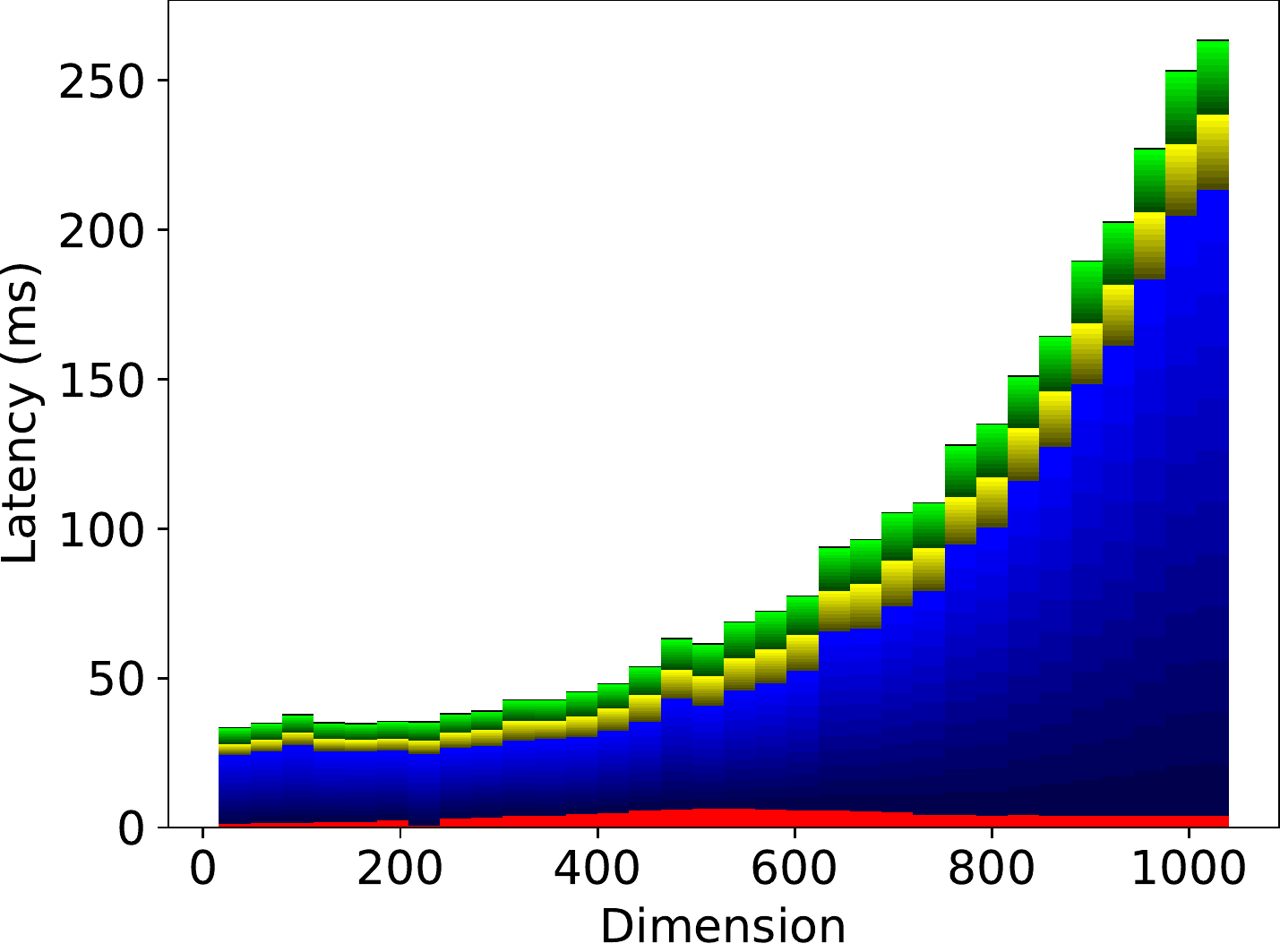}
    \\
    \end{tabular}\vspace{-1em}
    \captionof{figure}{Layerwise latency of different vanilla Transformer architectures in inference mode.}\vspace{-0.5em}
    \label{tab:layerwisevanilla}
\end{table*}

\begin{table*}[htb!]
    \centering
    \small
    \setlength{\tabcolsep}{1em}
    \begin{tabular}{c}
    \crule[blue]{0.8em}{0.8em} BERT \quad \crule[green]{0.8em}{0.8em} Nyströmformer \quad \crule[orange]{0.8em}{0.8em} Longformer \quad \crule[red]{0.8em}{0.8em} HuBERT \quad \crule[purple(html/css)]{0.8em}{0.8em} L-HuBERT \quad \crule[brown]{0.8em}{0.8em} ViT \quad \crule[softpink]{0.8em}{0.8em} Swin
    \end{tabular}
    
    \begin{tabular}{cc}
    \toprule
    Inference-Latency & Inference-Max-Memory\\
    \includegraphics[width=0.3\textwidth]{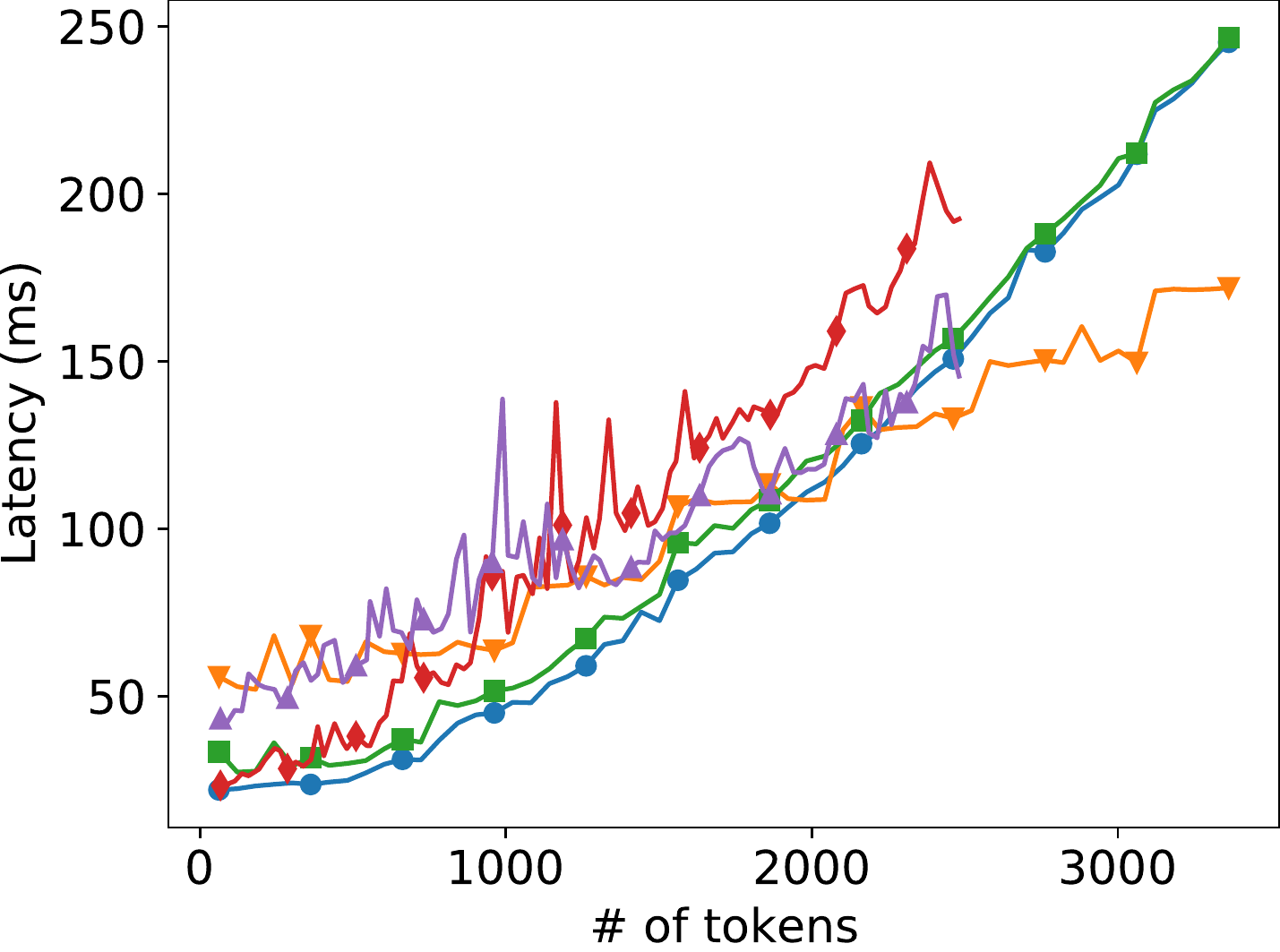} &
    \includegraphics[width=0.3\textwidth]{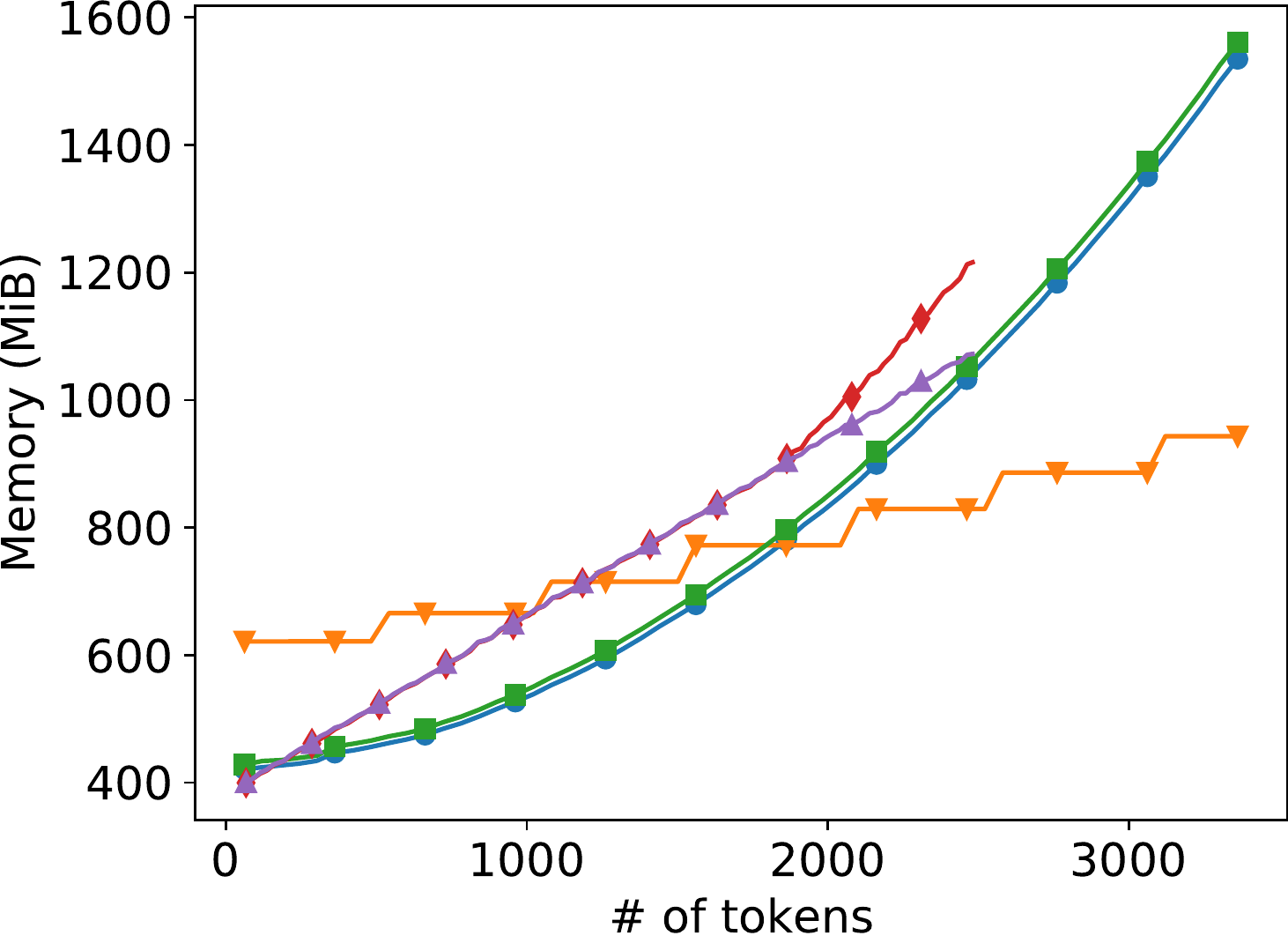} \\
    \includegraphics[width=0.3\textwidth]{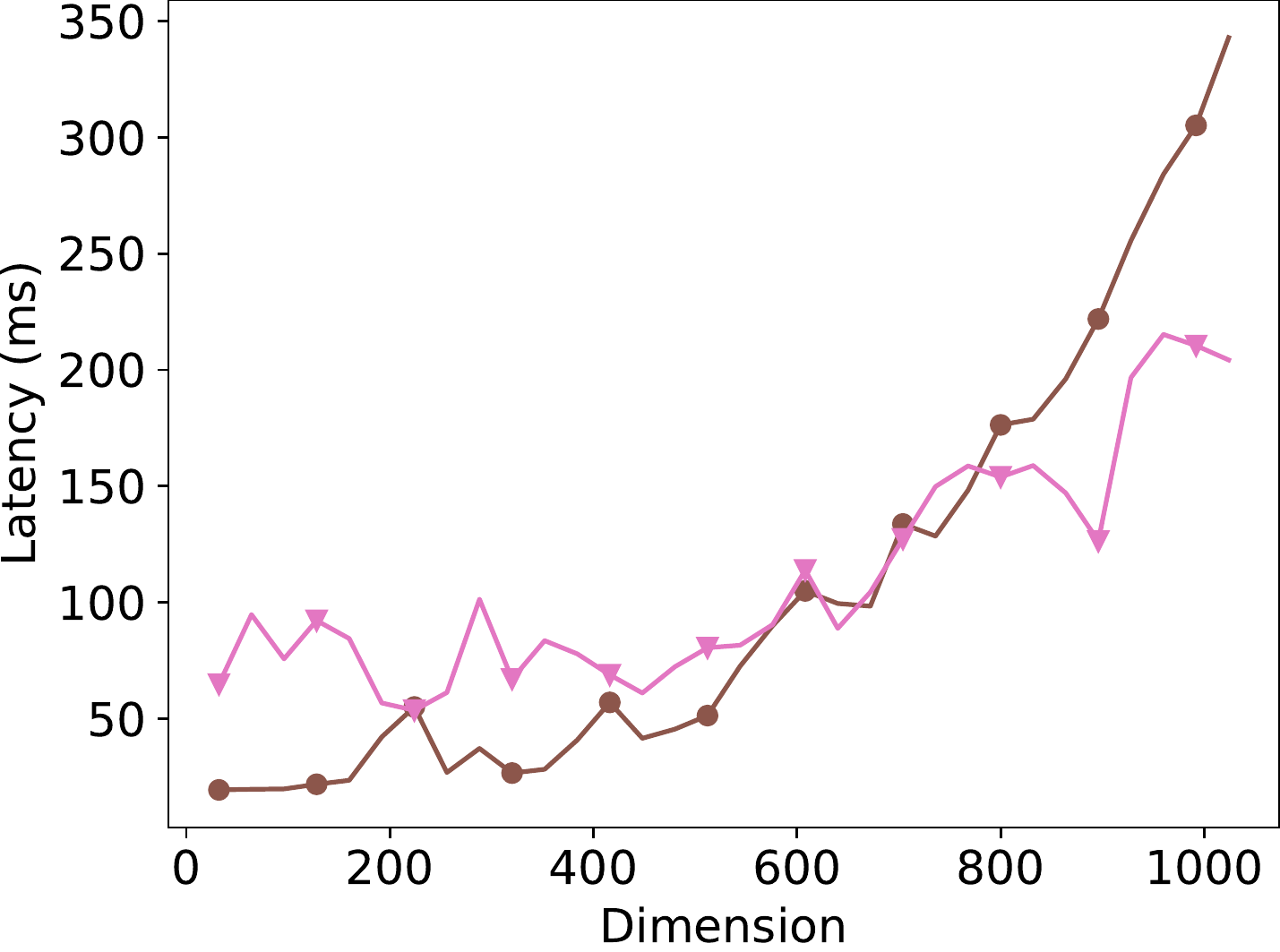} &
    \includegraphics[width=0.3\textwidth]{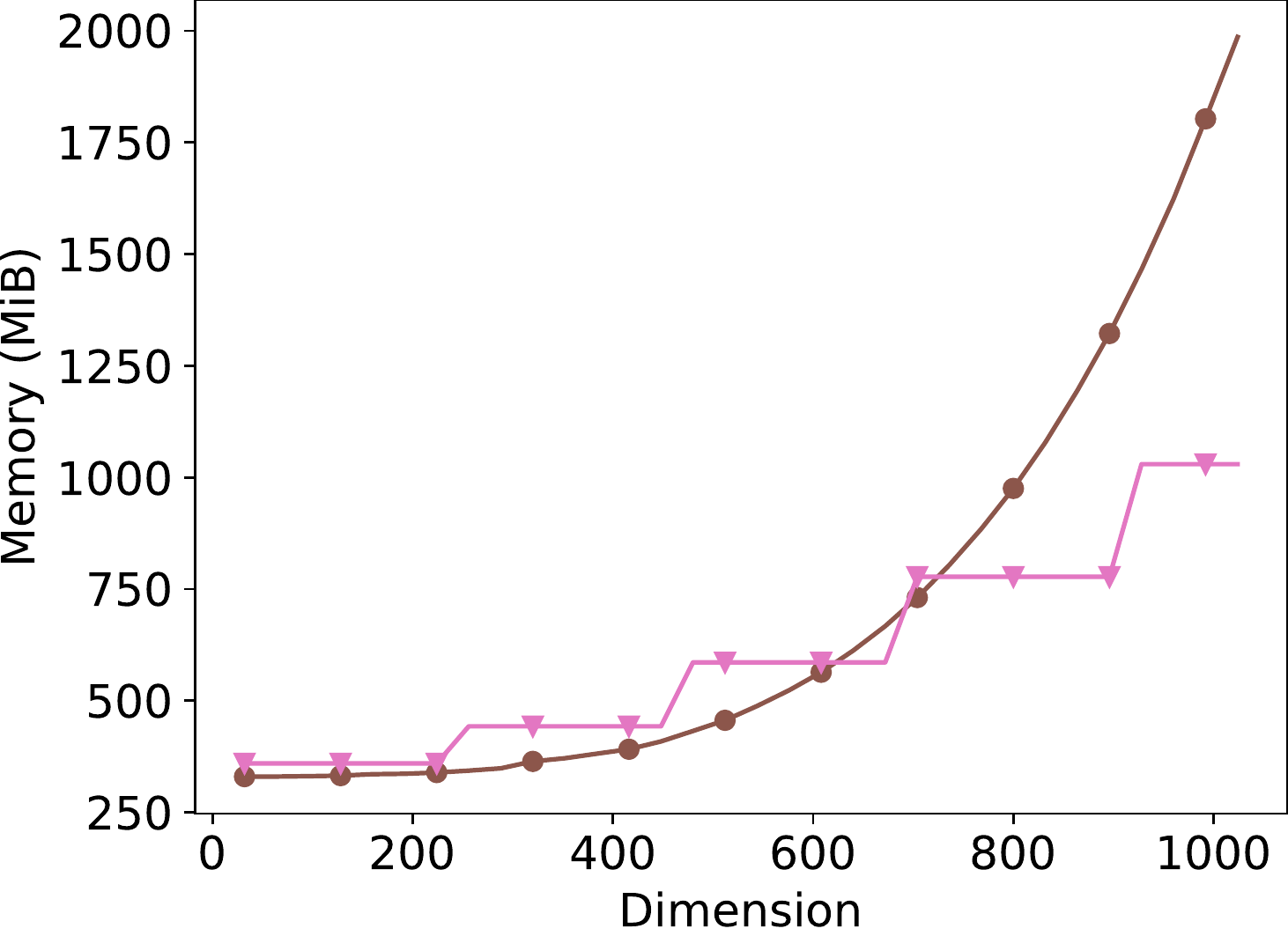}
    \end{tabular}
    \captionof{figure}{Overall Inference-time Profiling Results. Text and speech models in first row, vision models in second.}
    \label{tab:profilingresults_inference}
    \vspace{-12pt}
\end{table*}

\begin{table*}[htb!]
    \centering
    \small
    \setlength{\tabcolsep}{1em}
    \begin{tabular}{c}
    \crule[blue]{0.8em}{0.8em} BERT \quad \crule[green]{0.8em}{0.8em} Nyströmformer \quad \crule[orange]{0.8em}{0.8em} Longformer \quad \crule[red]{0.8em}{0.8em} HuBERT \quad \crule[purple(html/css)]{0.8em}{0.8em} L-HuBERT \quad \crule[brown]{0.8em}{0.8em} ViT \quad \crule[softpink]{0.8em}{0.8em} Swin
    \end{tabular}
    
    \begin{tabular}{cc}
    \toprule
    Training-Throughput & Training-Max-Memory\\
    \includegraphics[width=0.3\textwidth]{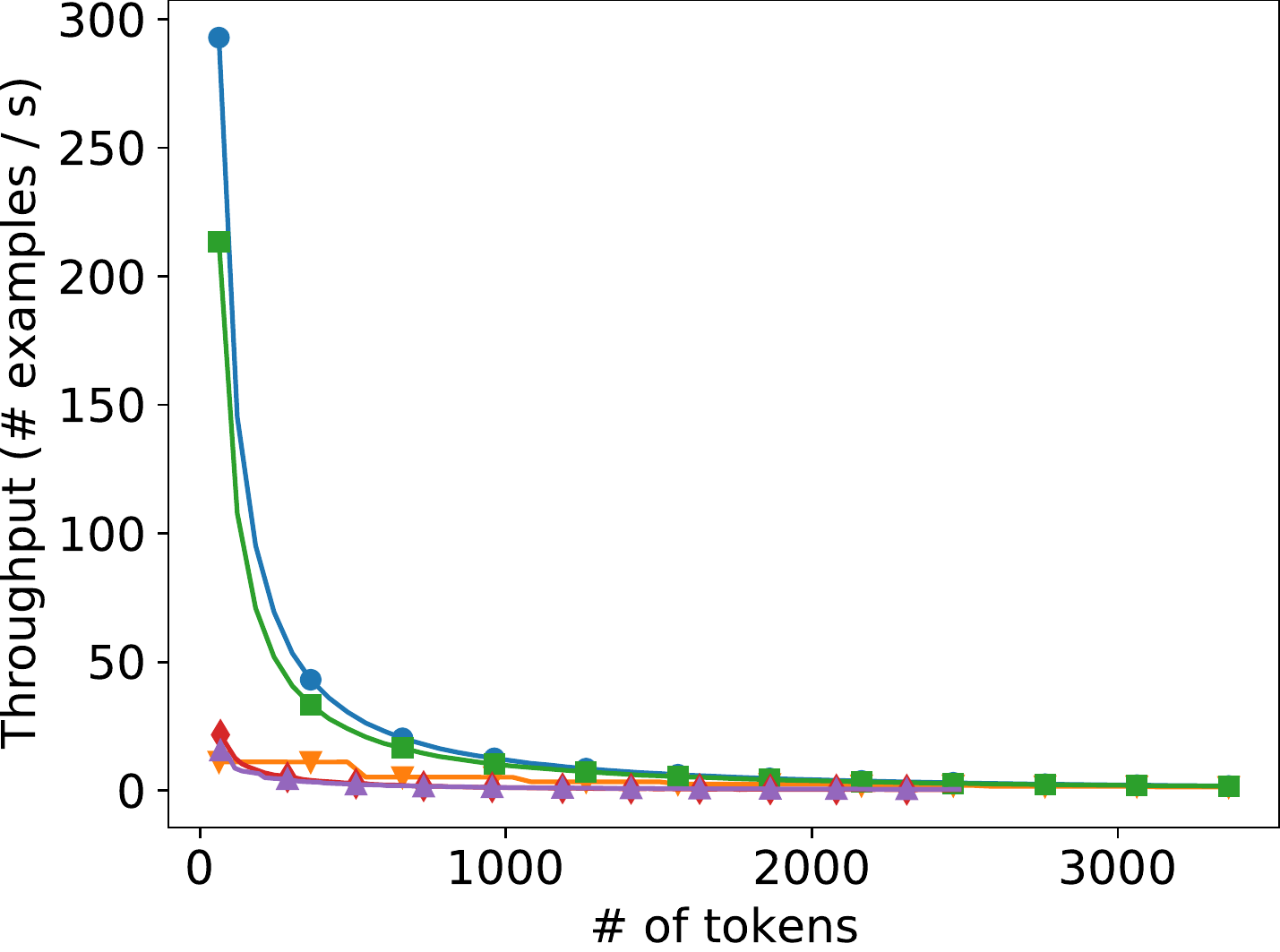} &
    \includegraphics[width=0.3\textwidth]{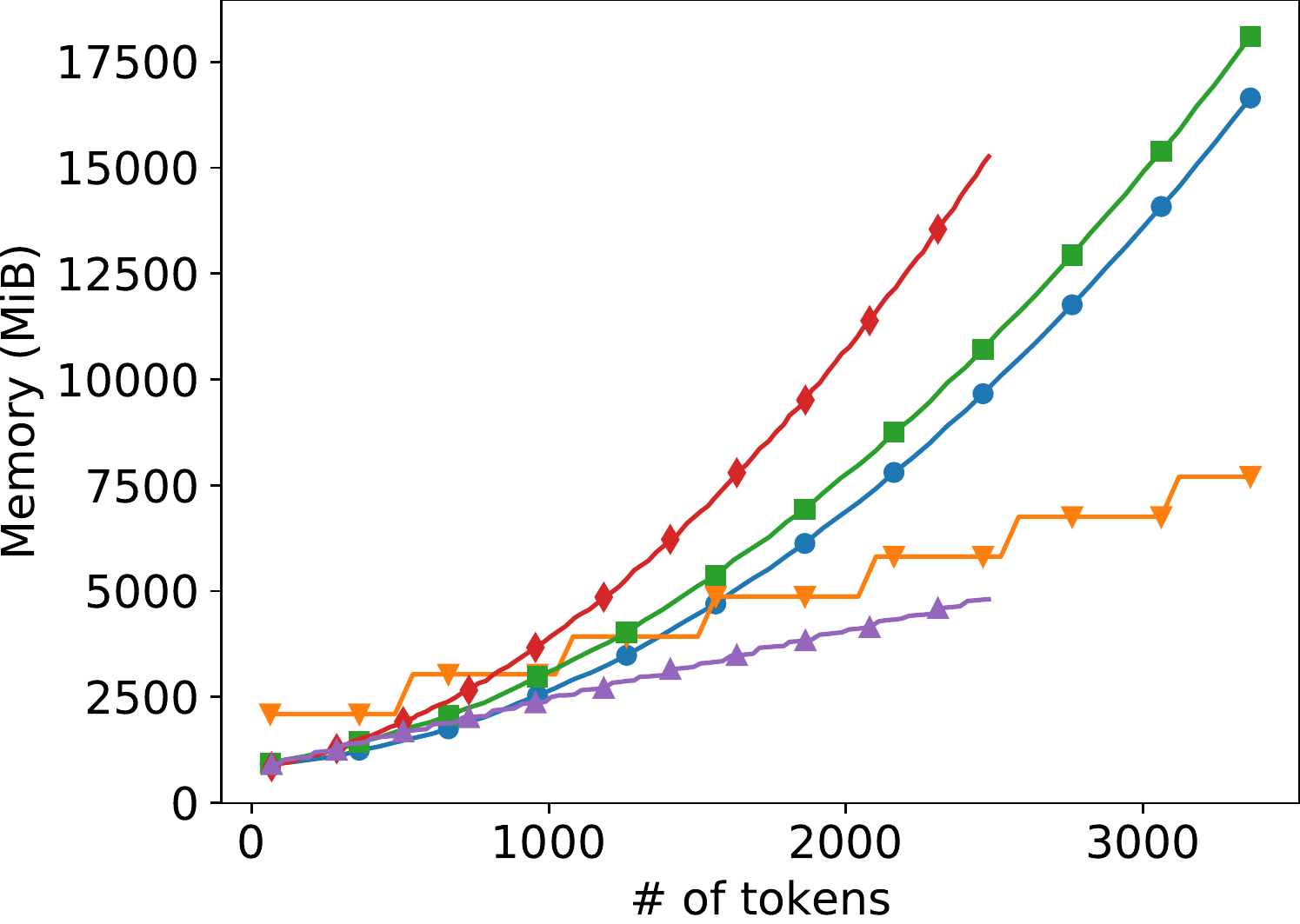} \\
    \includegraphics[width=0.3\textwidth]{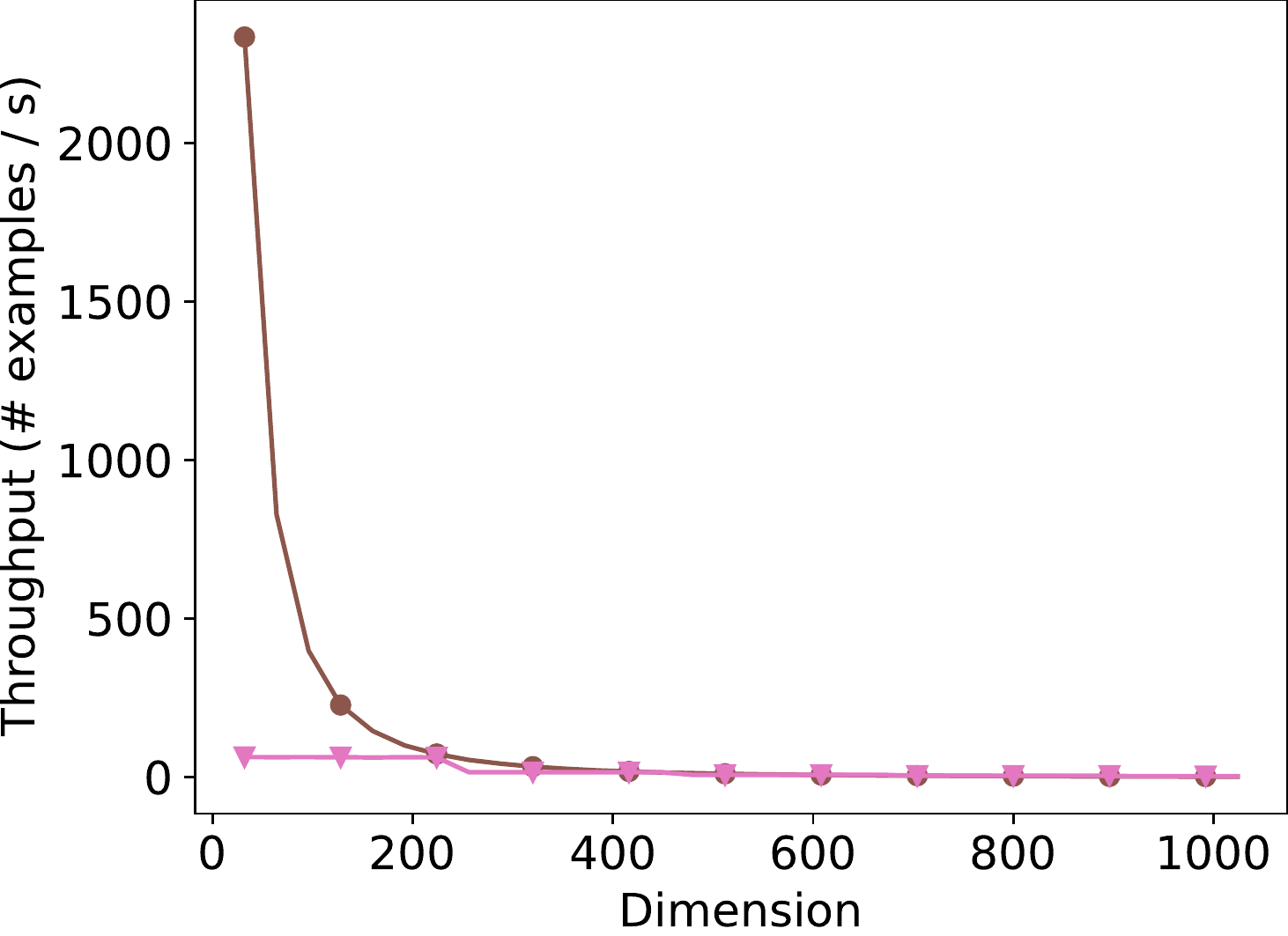} &
    \includegraphics[width=0.3\textwidth]{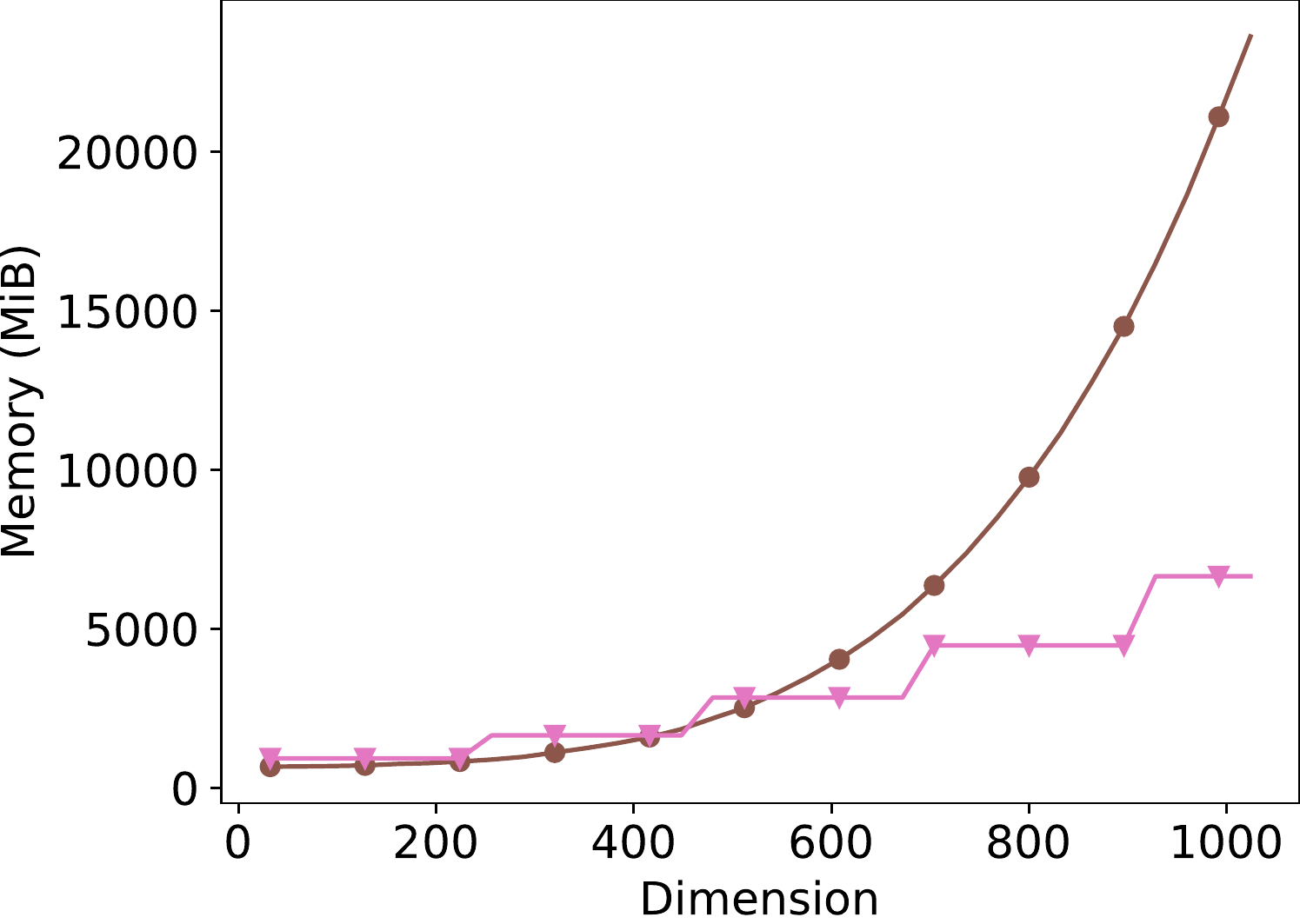}
    \end{tabular}
    \captionof{figure}{Overall Training-time Profiling Results. Text and speech models in first row, vision models in second.}
    \label{tab:profilingresults_training}
    \vspace{-12pt}
\end{table*}

Figure~\ref{tab:layerwisevanilla} shows the layerwise Latency-Inference for all 3 vanilla architectures in each modality. Figures for efficient models are in Appendix~\ref{app:other}. 
Color darkness represents the layer index (layer 0 is darkest). Table~\ref{tab:layer_wise_param} shows the layerwise param count.

Asymptotically, self-attention dominates the computation. However, since the average seq length for most text and speech tasks is less than $1000$ tokens and most image datasets are used at a max dimension of $512$, at these points, non-self-attention components take up $35\%$, $58.8\%$ and $43.75\%$ latency for NLP, speech and images. Additionally, parameter counts of SA are also comparable to Interm/Output layers. This shows that it is also important to direct efficiency efforts for other model components.

While the latency associated with embedding layers is minimal for BERT, they are sizable for HuBERT. HuBERT uses a CNN feature extractor with different strides and kernel sizes and consumes more time in the earlier CNN layers as opposed to later ones, as is visible in Figure~\ref{tab:layerwisevanilla}, which shows darker shades i.e. earlier layers dominating the computation. Optimal efficiency strategies can thus differ across modalities, e.g.~\citet{Wu2022PerformanceEfficiencyTI} slims down this CNN feature extractor embedding layer. On the other hand, embedding layers take up a lot of \textit{parameters} in BERT; thus, it may be helpful to shrink the BERT embedding layer for memory purposes (as opposed to \textit{latency} for HuBERT).
Finally, analyzing Transformer variants (Appendix~\ref{app:other}), we see that self-attention in Longformer, Swin and L-HuBERT encouragingly scales latency linearly, but with large overhead for smaller inputs.

\subsection{Overall Profiling Results}

Our profiling results are in Figures~\ref{tab:profilingresults_inference} and~\ref{tab:profilingresults_training}. Inference Throughput is in the Appendix at Figure~\ref{tab:throughputinferenceresults}, exhibiting similar trends as training Throughput.

 \paragraph{Tipping Point Analysis}  %
We see that most variants are slower and more memory hungry than vanilla models for input lengths of typical-context tasks. We define the \textit{tipping point} for each modality: the input length at which the variant becomes more efficient than the vanilla model. For text and speech, it is $1750-2000$ tokens for inference latency and max-memory, greater than typical input lengths (Table~\ref{tab:inputranges}). However, while the tipping point for training max-memory is $\approx 1500$ tokens for text (still a large number), it is $\approx 0-250$ for speech, an encouraging result. For images, it is $500-700$ pixels for all metrics apart from throughput. This is less reasonable for $224$ pixel datasets but good for high resolution image datasets ($512/1024$). All variants are either worse or comparable than vanilla models across modalities for throughput. 

We hypothesize that some efficient models suffer from additional overheads; while vanilla attention benefits from highly optimized matrix multiplication, windowed attention requires complex reshaping and preprocessing. 

\paragraph{Choosing the Right Model Depends on Resource Constraints} 
Our results show that the choice of the right model depends on resource constraints. Suppose that one is training models under a time constraint; then, throughput is the bottleneck and efficient models would not be a good fit. On the other hand, efficient models are useful for long context memory-constrained inference.

\paragraph{Local Attention and Excessive Padding}
The Longformer pads input lengths to be a multiple of $512$ and Swin requires input dimension to be a multiple of $224$. This slows shorter inputs down and results in extremely low performance (measured by all 3 metrics) as compared to vanilla models.

\paragraph{Comparing Parameter Counts}
The Longformer uses more parameters compared to vanilla BERT (148M vs. 109M) because it uses two sets of Q,K,V projection matrices for its global and local attention operations; sharing these may decrease its memory usage. For other modalities, efficient models do not incur more parameters.

\section{Conclusion}
We present an empirical efficiency analysis of vanilla Transformers and their self-attention-based efficient variants across modalities, metrics and input context sizes. We find substantial differences across modalities and metrics when analyzing the tipping point for efficient variants. Finally, the layerwise analysis finds that self-attention is not the only bottleneck. We recommend that all efficient model papers should report such cross-modal, layerwise profiling results on multiple efficiency metrics covering a variety of use-cases to provide a full picture of the benefits of the model.

\todo[inline]{\scriptsize{anuj: Commented out the 'For practitioners choosing between models' section because rereading this part, I'm not sure if we actually say anything interesting here. We're just talking about standard things most people know; it's not related to our expts.}}

\section*{Limitations}
We focus primarily on comparing model efficiencies using a variety of efficiency metrics and do not consider model performance; one can perform a more elaborate analysis of performance-efficiency tradeoffs, which we did not do here.

We only profile a total of seven models across three modalities while there are more efficient variants and vanilla Transformers proposed in the literature. While we choose our models to be as representative of each modality and efficiency technique as possible, we cannot extrapolate results to other model variants and other modalities. In particular, modalities like video and genomics and efficiency approaches like quantization would be interesting to profile, which we did not do.

 \section*{Acknowledgements}
 We thank the reviewers and the meta-reviewer of the ACL community for helpful feedback on the draft. This work was partially funded by a grant from UT Machine Learning Lab.

\bibliography{anthology,custom}
\bibliographystyle{acl_natbib}

\appendix

\section{Sampling Speech Utterances for Profiling}
\label{app:speechinputrange}
To obtain speech inputs of length $i$ seconds to $i + 0.5$ seconds for all $i$ less than $12$ seconds, we sample $5$ speech utterances from the training set of the Librispeech dataset~\cite{librispeech} whose lengths fall within this range and compute aggregate metrics over these $5$ utterances. Since the Librispeech dataset does not contain extremely long speech utterances, for $i$ of length greater than $12$ seconds, we adopt a different approach to generate inputs. To generate such an input utterance of length between $i$ and $i+0.5$ seconds, we first sample $5$ speech utterances from the Librispeech training set of input length ranging from $\frac{i}{5}$ to $\frac{i+0.5}{5}$ and concatenate them to obtain utterances of length ranging from $i$ to $i+0.5$ as desired. We do this $5$ times to get $5$ different utterances and compute aggregate metrics over these $5$ utterances.

\section{Computing Token Lengths for NLP and Speech Datasets}
\label{app:tokenlengths}
We compute average sequence token lengths for $7$ NLP datasets and $6$ speech datasets. For all speech datasets, we compute mean utterance durations and multiply durations by $50$ to obtain number of tokens (model framerates are $20$ ms i.e. $\times 50$). For TEDLIUM~\cite{Hernandez_2018}, LJSpeech~\cite{ljspeech17}, VoxCeleb Speaker Recognition Dataset~\cite{nagrani17_interspeech} and Librispeech~\cite{librispeech}, we compute mean validation-set \textit{utterance} durations; for Spoken SQuAD~\cite{li2018spoken}, we report mean validation-set \textit{paragraph} duration and for the Spotify English Podcasts dataset~\cite{clifton-etal-2020}, we report mean \textit{podcast} duration directly obtained from~\citet{clifton-etal-2020}.

\noindent \textbf{SST}~\cite{socher-etal-2013-recursive}. We use test-set sentences. We use the HuggingFace BERTTokenizer.

\noindent \textbf{MNLI}~\cite{williams-etal-2018-broad}. We use validation-matched-set examples by concatenating the premise and the hypothesis. We use the HuggingFace BERTTokenizer.

\noindent \textbf{SQuAD2.0}~\cite{rajpurkar-etal-2018-know}. We use validation-set examples by concatenating the context and the question. We use the HuggingFace BERTTokenizer.

\noindent \textbf{OntoNotes}~\cite{pradhan-xue-2009-ontonotes}. We obtain this number from the Longformer~\cite{Beltagy2020Longformer} paper.

\noindent \textbf{CNN-Dailymail}~\cite{NIPS2015_afdec700}. We use the 3.0.0 version of the dataset and use test-set articles. We use the HuggingFace BERTTokenizer.

\noindent \textbf{HotpotQA}~\cite{yang-etal-2018-hotpotqa}. We obtain this number from the Longformer~\cite{Beltagy2020Longformer} paper.

\noindent \textbf{TriviaQA}~\cite{joshi-etal-2017-triviaqa}. We obtain this number from the Longformer~\cite{Beltagy2020Longformer} paper.

\section{Implementing Throughput Profiling}
\label{app:throughput}
To profile Throughput, we need to compute the max batch size that can fit on a single GPU. We \textit{approximately} predict this using a linear estimator as follows. We first record the memory $B$ reserved on the GPU after just loading the model. Next, we independently run batches of sizes $1$ and $2$ and record memory usages $M_1$ and $M_2$. We use an NVIDIA Quadro RTX 8000 GPU with a maximum memory of $45000$ MiB. Thus, assuming a linear relationship between batch size and memory consumption, we predict a maximum batch size of $bsz=\frac{45000-B}{M_2-M_1}$. In practice, this is an overestimate; we keep decreasing the batch size by a factor of $0.9$ until no OOM errors occur and this is our final estimate.

\section{Implementational Details for Models}
\label{app:implmodels}
We use the following HuggingFace configurations: \texttt{bert-base-uncased} for BERT, \texttt{allenai/longformer-base-4096} for Longformer, \texttt{uw-madison/nystromformer-4096} for Nyströmformer, \texttt{google/vit-base-patch16-224} for ViT and \texttt{microsoft/swin-base-patch4-window7-224} for Swin. The BERT model natively supports a maximum of $512$ tokens as input because it has $512$ positional embeddings; we modify the positional embedding computation to allow an arbitrarily long input to be provided. The Longformer internally pads all input lengths to a multiple of $512$. For Swin, we pad images to have an input dimension that is a multiple of $224$; this is necessary due to the windowed attention mechanism in Swin. In fact, the Swin model natively supports only a $224 \times 224$ resolution; we make a small modification in order to support resolutions that are multiples of $224$. We use the \texttt{HuBERT Base} model for both HuBERT and L-HuBERT.

\section{Transformer Layer Types}
\label{app:layerdescriptions}
\noindent \textbf{Input Embedding Layer.} (\crule[red]{0.8em}{0.8em}/red) Maps the input sequence into fixed-dimensional embeddings. This is a linear layer for text and a CNN featurizer for image/speech.

\noindent \textbf{Positional Embedding Layer.} (\crule[fuchsia]{0.8em}{0.8em}/fuchsia) For text and image models this is part of the input embedding layer. For speech models, this is a very wide convolution layer.

\noindent \textbf{Self Attention Layer.}(\crule[blue]{0.8em}{0.8em}/blue) The multi-head self attention block, which computes self-attention outputs and maps the result to the model dimension.

\noindent \textbf{Intermediate Layer.}(\crule[yellow]{0.8em}{0.8em}/yellow) Linear layer of the feedforward block that maps the output from the Self Attention block into the `feedforward dimension' (typically 4x the model dimension).

\noindent \textbf{Output Layer.}(\crule[green]{0.8em}{0.8em}/green) Second linear layer of the feedforward block, which maps the output from Intermediate layer back to the model dimension.

\noindent \textbf{Other Layers.}(\crule[black]{0.8em}{0.8em}/black) Other modules (activations, layer normalizations, other linear layers, etc.) not covered by the above components.

\section{Additional Profiling Analyses}
\label{app:other}
We report layerwise profiling runs for efficient self-attention variants and inference-time throughput profiling runs for all variants in this section at Figures~\ref{tab:layerwiselatencyvariants} and~\ref{tab:throughputinferenceresults}.
\begin{table*}
    \small
    \centering
    \begin{tabular}{c}
    \crule[red]{0.8em}{0.8em} Input Embedding \quad \crule[fuchsia]{0.8em}{0.8em} Positional Embedding \quad \crule[blue]{0.8em}{0.8em} Self-Attention \quad \crule[yellow]{0.8em}{0.8em} Intermediate \quad \crule[green]{0.8em}{0.8em} Output \quad 
    \crule[black]{0.8em}{0.8em} Other \\
    \end{tabular}
    \vspace{1em}
    \begin{tabular}{cc}
    Longformer & Nyströmformer \\
         \includegraphics[width=0.4\textwidth]{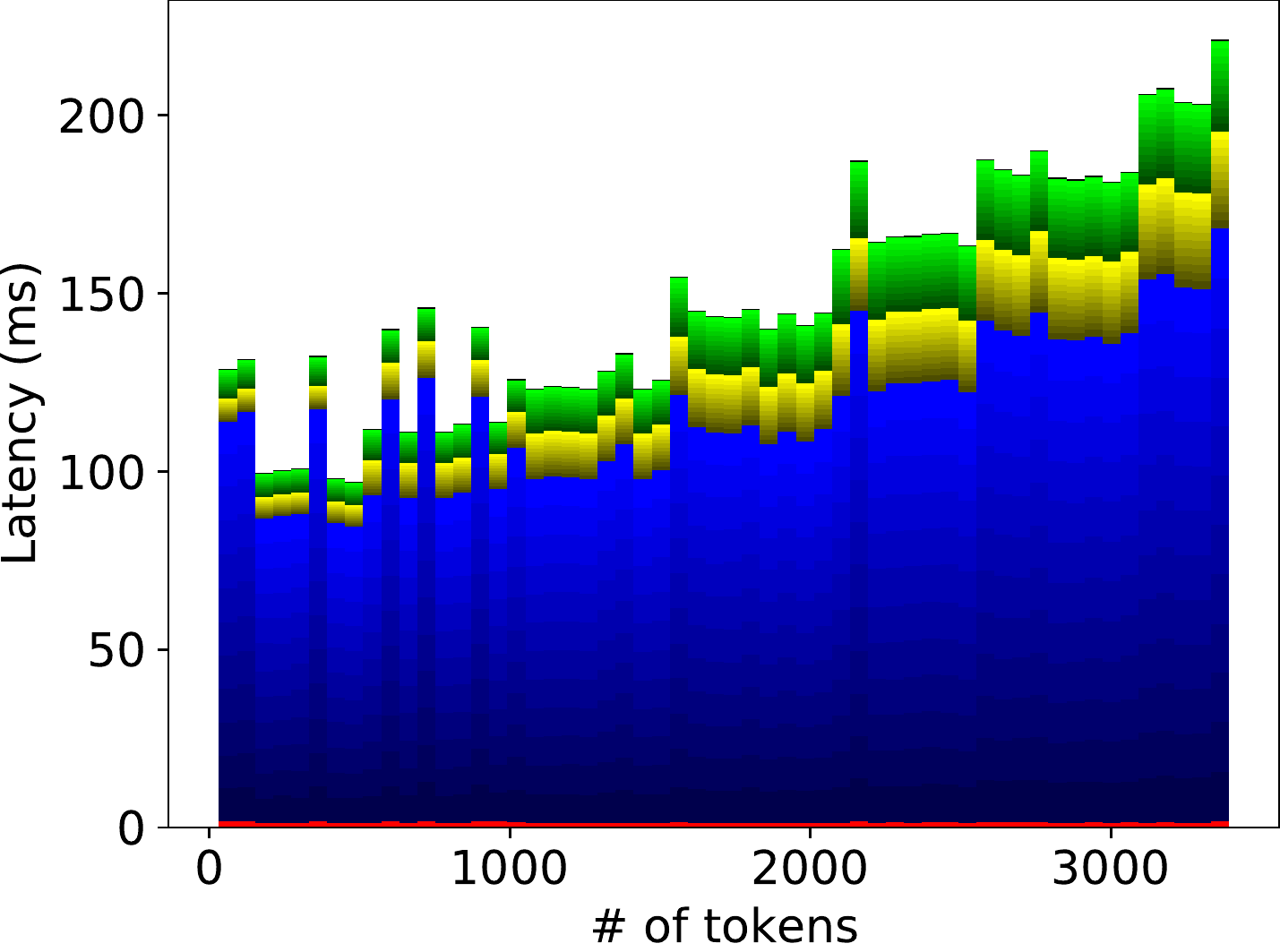} &
        \includegraphics[width=0.4\textwidth]{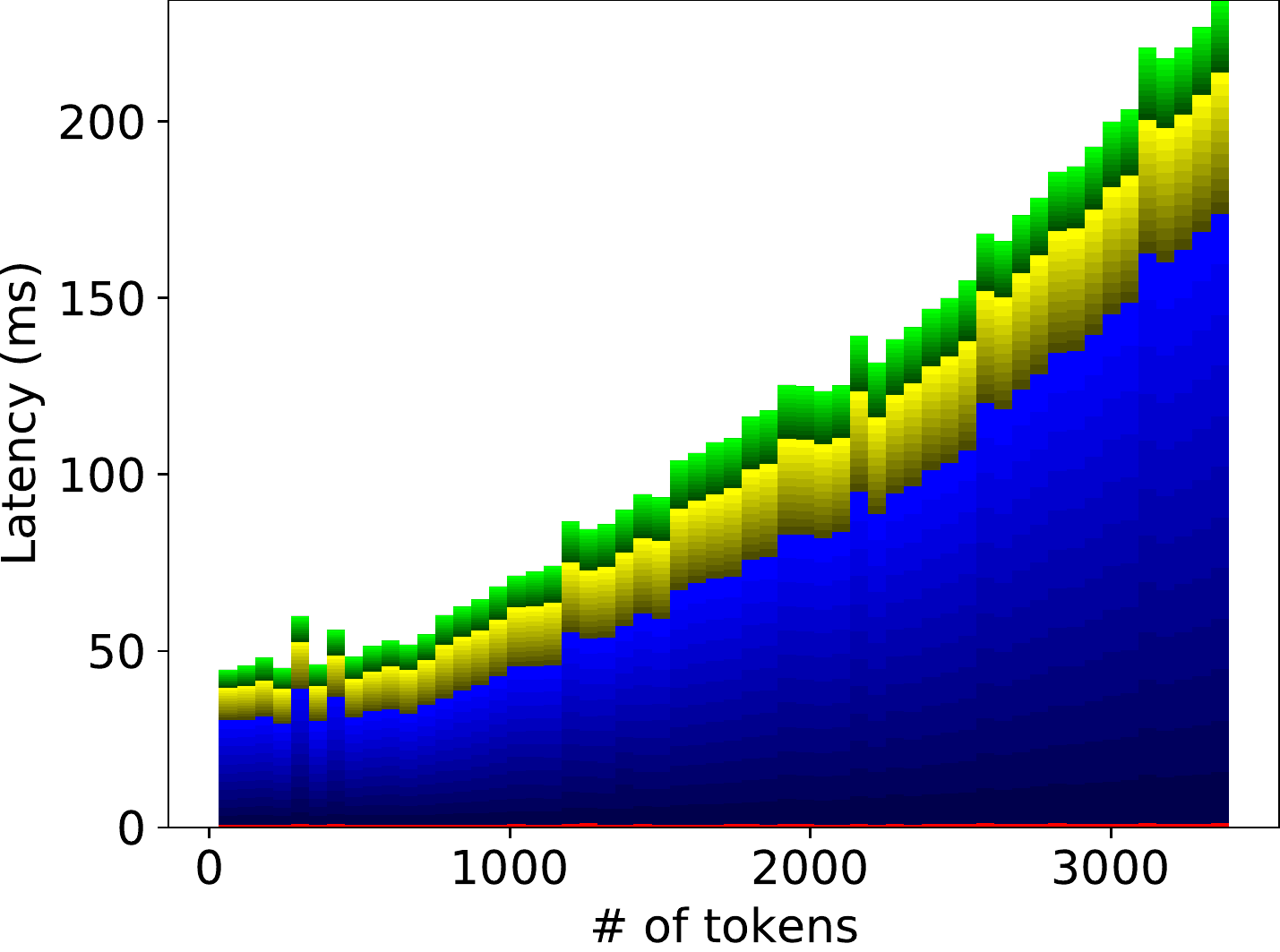}         \vspace{1em} \\
         L-HuBERT & Swin \\
        \includegraphics[width=0.4\textwidth]{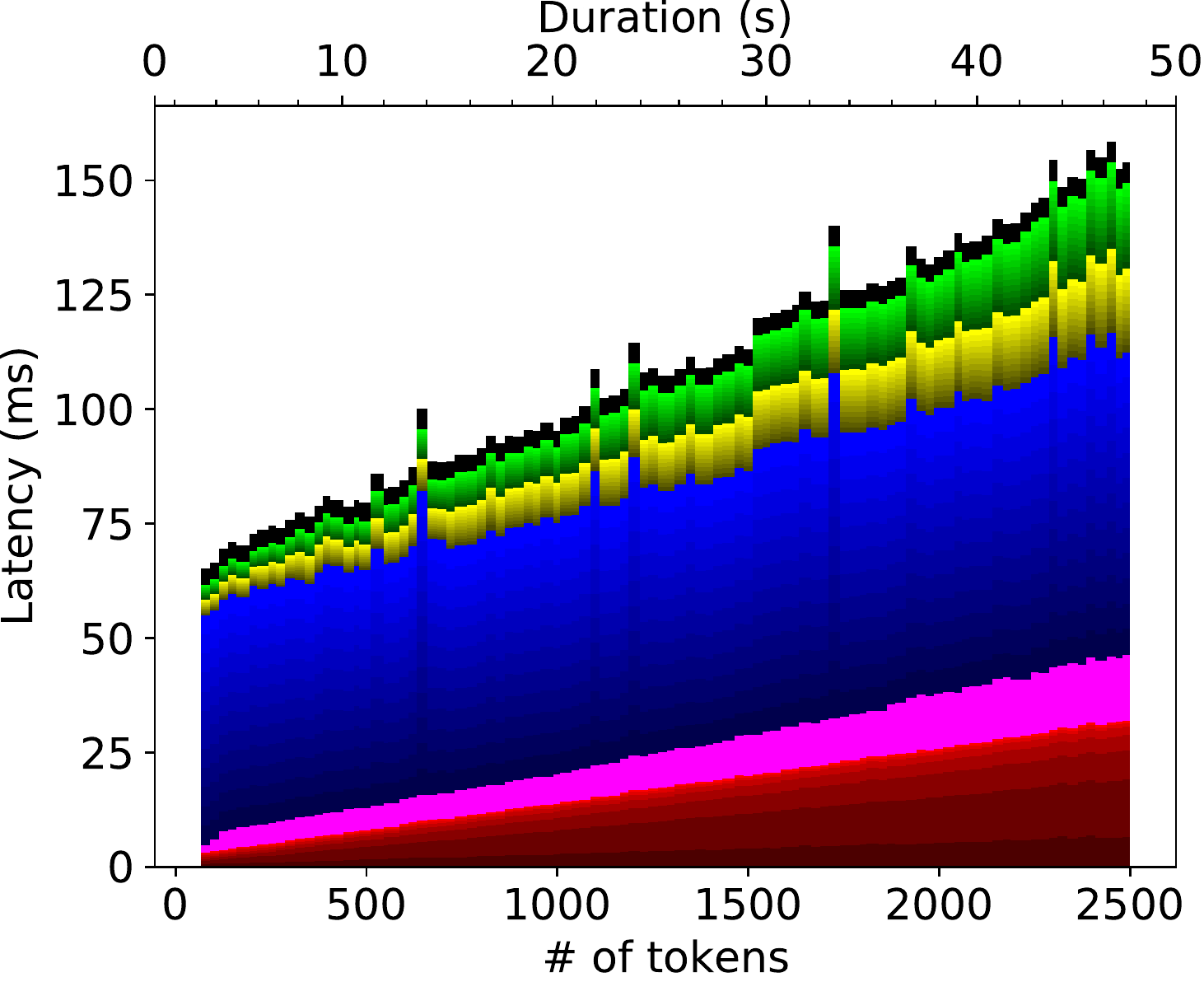} &
        \includegraphics[width=0.4\textwidth]{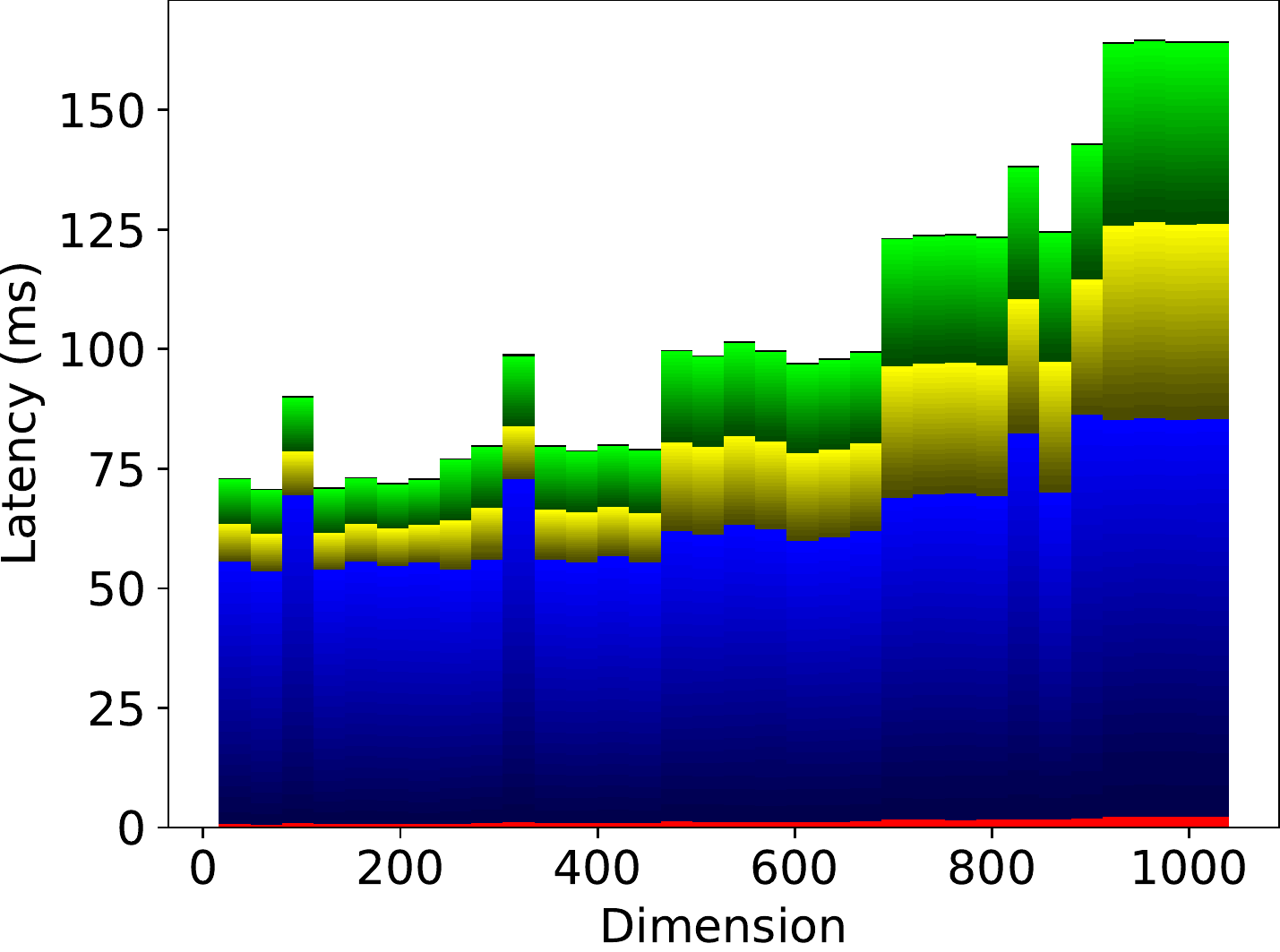}
    \end{tabular}
    \captionof{figure}{Layerwise latency of different Transformer variants in inference mode.}
    \label{tab:layerwiselatencyvariants}
\end{table*}

\begin{table*}
    \centering
    \small
    \begin{tabular}{cc}
    Text+Speech & Image \\
    \includegraphics[width=0.4\textwidth]{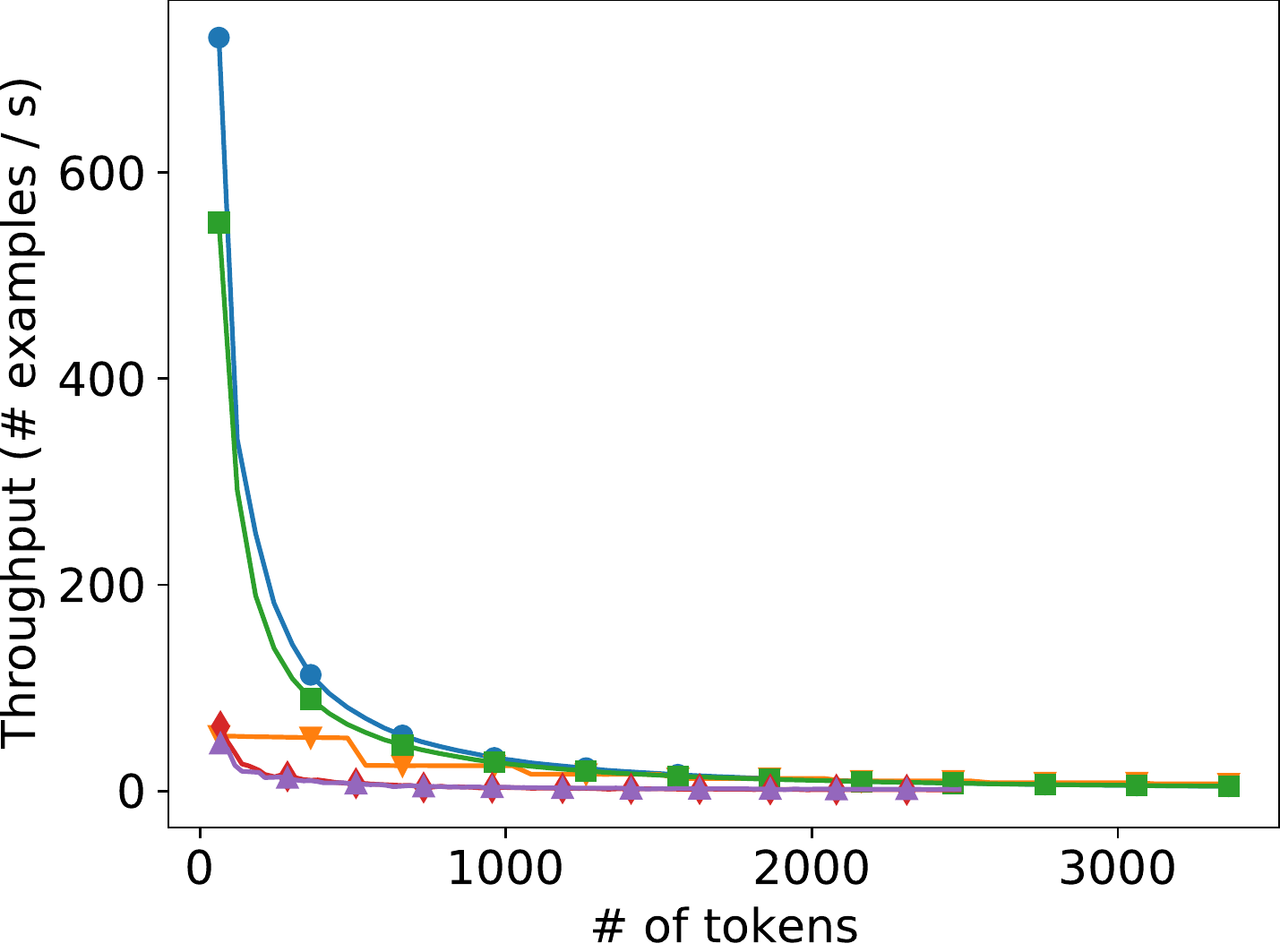} & \includegraphics[width=0.4\textwidth]{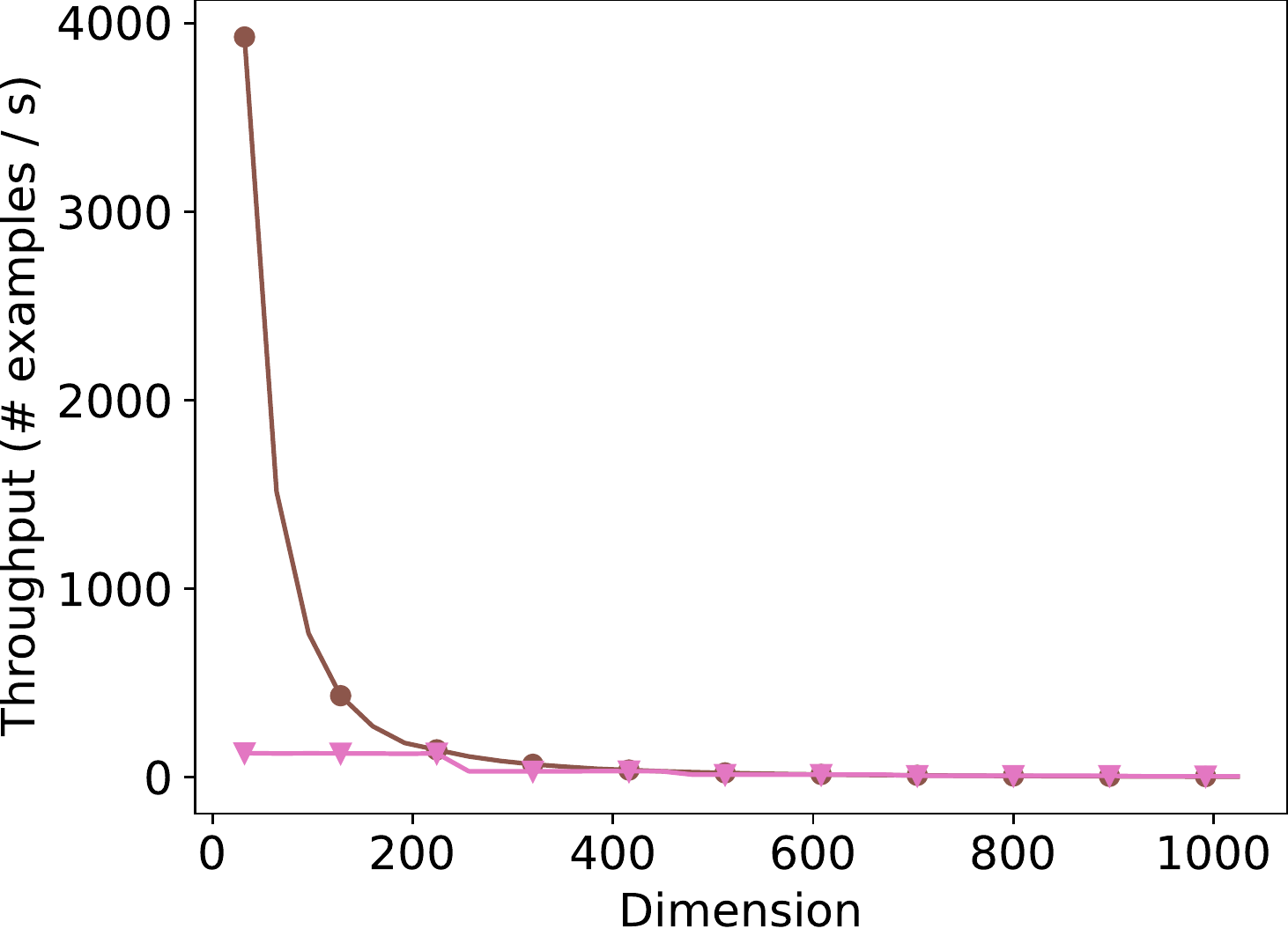} \\
    \end{tabular}
    \captionof{figure}{\textbf{Throughput} Profiling Results in inference mode.}
    \label{tab:throughputinferenceresults}
\end{table*}

\end{document}